\documentclass[journal]{IEEEtran}

\usepackage{amsmath,amsfonts,bm}
\usepackage{array}
\usepackage{textcomp}
\usepackage{stfloats}
\usepackage{url}
\usepackage{verbatim}
\usepackage{graphicx}
\usepackage{cite}
\usepackage{amssymb}
\usepackage{CJK}
\usepackage[ruled,linesnumbered, noend]{algorithm2e}
\usepackage{booktabs}
\usepackage{diagbox}

\ifCLASSOPTIONcompsoc
\usepackage[caption=false, font=footnotesize, labelfont=sf, textfont=sf,subrefformat=parens]{subfig}
\else
\usepackage[caption=false, font=footnotesize,,subrefformat=parens]{subfig}
\usepackage{makecell}
\usepackage{multirow}
\usepackage{xcolor}
\usepackage{tabularx}
\usepackage[table]{xcolor} % 在导言区加载包
\usepackage{bbding}
\usepackage{threeparttable}
\usepackage{color}
\usepackage{mathtools}
\usepackage{ulem}
\usepackage{pifont}
\usepackage{threeparttable}
\usepackage{ragged2e}  % 提供 \justifying 命令

\definecolor{HeaderGray}{HTML}{A9A9A9}   % 第一行：深灰
\definecolor{FirstColGray}{HTML}{E6E6E6} % 第一列：浅灰
\definecolor{RowColorA}{HTML}{DCEAF7}    % 奇数行颜色
\definecolor{RowColorB}{HTML}{EFF9EA}    % 偶数行颜色

\newcolumntype{X}{>{\RaggedRight\arraybackslash}p{5cm}}

\begin{document}

\title{Agentic AI for ISAC: Analysis, Framework, and Case Study}

\author{
        Wenwen Xie,
        Geng Sun,~\IEEEmembership{Senior Member,~IEEE,}
        Chuang Zhang,
        Xuejie Liu,
        Dong In Kim,~\IEEEmembership{Fellow,~IEEE}
        \thanks
        {
        % \par This study is supported in part by the National Natural Science Foundation of China (62272194, 62471200), and in part by the Science and Technology Development Plan Project of Jilin Province (20250101027JJ).
        \par Wenwen~Xie, Geng~Sun, Chuang~Zhang, and Xuejie~Liu are with the College of Computer Science and Technology, Jilin University, Changchun 130012, China~(e-mails: xieww22@mails.jlu.edu.cn, sungeng@jlu.edu.cn, chuangzhang1999@gmail.com, xuejie@jlu.edu.cn).
        \par Dong In Kim is with the Department of Electrical and Computer Engineering, Sungkyunkwan University, Suwon 16419, South Korea (e-mail:dongin@skku.edu).

        \par (\textit{Corresponding author: Geng Sun.})
        }
 }

\maketitle

\begin{abstract}
Integrated sensing and communication (ISAC) has emerged as a key development direction in the sixth-generation (6G) era, which provides essential support for the collaborative sensing and communication of future intelligent networks. However, as wireless environments become increasingly dynamic and complex, ISAC systems require more intelligent processing and more autonomous operation to maintain efficiency and adaptability. Meanwhile, agentic artificial intelligence (AI) offers a feasible solution to address these challenges by enabling continuous perception-reasoning-action loops in dynamic environments to support intelligent, autonomous, and efficient operation for ISAC systems. As such, we delve into the application value and prospects of agentic AI in ISAC systems in this work. Firstly, we provide a comprehensive review of agentic AI and ISAC systems to demonstrate their key characteristics. Secondly, we show several common optimization approaches for ISAC systems and highlight the significant advantages of generative artificial intelligence (GenAI)-based agentic AI. Thirdly, we propose a novel agentic ISAC framework and prensent a case study to verify its superiority in optimizing ISAC performance. Finally, we clarify future research directions for agentic AI-based ISAC systems.
\end{abstract}


%section
%Introduction
%
\section{Introduction}
\par With the rapid evolution of communication technologies and the widespread deployment of radar systems, wireless spectrum resources have become increasingly scarce. In this case, integrated sensing and communication (ISAC) has emerged to improve spectrum efficiency and has been considered one of the most promising technologies in the sixth-generation (6G) era~\cite{Qaisar2026}. By integrating communication and sensing within a unified framework, 6G systems can support both high-quality communication and high-resolution sensing. According to recent forecasts, the global ISAC market reached 3.54 billion in 2024 and is expected to grow to 12.5 billion by 2035\footnote{https://www.wiseguyreports.com/reports/integrated-sensing-and-communication-isac-market}. This trend reflects the growing demand for intelligent ISAC systems. Therefore, it is essential to identify effective approaches for advancing ISAC development.

\par However, conventional artificial intelligence (AI) approaches alone are insufficient to address the complex decision-making requirements of future ISAC systems in highly dynamic environments~\cite{Cui2025}. For example, although deep reinforcement learning (DRL) has been widely applied in ISAC systems, its limited generalization capability restricts adaptation to changing tasks and wireless environments. Moreover, although large language models (LLMs) possess strong generative and reasoning capabilities, hallucinations may reduce decision reliability. Therefore, a new AI paradigm is required to support the evolution of ISAC systems.

\par Agentic AI offers a promising solution to these challenges. Built upon the perception-reasoning-action loop, agentic AI can perform explicit reasoning and objective-driven decision-making by combining environmental feedback with accumulated knowledge~\cite{Zhang2025}. Different from conventional AI methods, agentic AI integrates multiple advanced generative frameworks and external tools, enabling autonomous planning, task decomposition, and task execution. As a result, agentic AI can achieve adaptation and reasoning in complex environments. Recently, agentic AI has demonstrated remarkable effectiveness in various domains. For example, AutoGPT\footnote{https://github.com/Significant-Gravitas/AutoGPT}, a representative agentic AI system based on state-of-the-art LLMs (such as GPT-4), can autonomously decompose complex tasks and interact with external software or online services. Similarly, agentic AI has shown significant potential for improving vehicle autonomy and adaptability in autonomous driving~\cite{Sapkota2026}. 
% For instance, it can formulate the joint sensing-communication process as a dynamic decision-making problem. By continuously perceiving the channel state information, target localization errors, and spectrum occupancy, agentic AI performs explicit reasoning based on historical experiences and environmental feedback to autonomously design beamforming, power allocation, and resource scheduling, \textit{etc}~\cite{Wang2025}. As such, agentic AI enables a real-time trade-off between communication quality and sensing accuracy of ISAC systems.

\par These examples highlight the importance of agentic AI for developing intelligent and autonomous systems. Inspired by this, agentic AI can be employed to address complex optimization problems in ISAC systems. To achieve this, several key issues should be further discussed. First, it is important to identify which ISAC optimization problems have to be effectively addressed. Second, we need to explore how effective agentic AI is in addressing these problems and how to exploit its potential to further improve both sensing and communication performance. To this end, we provide a forward-looking perspective on agentic AI for decision-making in ISAC systems. The contributions of this work are summarized as follows:
\begin{itemize}
    \item We first provide a comprehensive overview of the advantages and evolution of agentic AI. Subsequently, we present the characteristics of different ISAC systems and illustrate the applications of agentic AI in several ISAC scenarios. This systematic overview not only clarifies the theoretical foundations but also offers crucial insights into how agentic AI can potentially revolutionize ISAC.
    \item We review several common optimization approaches and highlight the unique advantages of agentic AI in ISAC optimization. Moreover, we focus on the emerging generative AI (GenAI)-driven agentic AI and elaborate on how the generative capabilities of GenAI empower agentic ISAC systems, which is significant for further improving the performance of future adaptive and intelligent agentic AI-based ISAC systems. 
    \item We propose an agentic ISAC framework that integrates the DRL, GenAI, and LLM. Moreover, we present a case study to validate the proposed framework. Simulation results demonstrate that our solution achieves a significant performance improvement over conventional approaches. Notably, the proposed framework offers considerable versatility, which can be potentially adapted to diverse dynamic communication scenarios and complex system optimization challenges due to its powerful adaptive learning capability.
\end{itemize}

\section{Overview of Agentic AI and ISAC}
\label{Overview of Agentic AI and ISAC}
\par In this section, we provide a detailed introduction of agentic AI and ISAC.

\subsection{Concept and Evolution of Agentic AI}
\label{sec:Concept and evolution of agentic AI}
\par In this part, we present a comprehensive review of the basic concepts and main evolution.

\subsubsection{Definition of Agentic AI}
\par Conventional AI methods typically rely on prior assumptions (\textit{e.g.}, prior information about the system model and data distribution), and they are mainly proficient in executing predefined tasks. In contrast, agentic AI, based on the \textit{continuous perception-reasoning-action loop}, is capable of autonomous context interpretation, explicit reasoning, goal-driven decision-making, and feedback-based policy improvement with minimal or no human assistance, especially allowing the agentic AI system to handle complex tasks by decomposing and performing them independently and efficiently~\cite{Zhang2025}. As such, the characteristics of agentic AI are presented as follows. 
\begin{itemize}
    \item \textit{Autonomy}: Agentic AI can independently perform reasoning, make decisions, and proactively initiate actions without continuous human assistance. 
    % This autonomous capability enables agentic AI to maintain coherent operation even in the absence or limited presence of external guidance.
    \item \textit{Memory and Adaptability}: Agentic AI is capable of extracting and retaining critical information from previous interactions and learning processes. By utilizing the historical knowledge, agentic AI informs current decision-making to improve decision accuracy and adaptability. 
    % This continuous integration of past and present knowledge forms the foundation for more context-aware and resilient intelligence.
    \item \textit{Explicit Reasoning and Agent Coordination}: Agentic AI possesses the explicit reasoning capability, which enhances the transparency and interpretability of its decision-making process. Moreover, agentic AI typically consists of specialized agents designed for different tasks, and it can flexibly coordinate these agents to process the reasoning results while invoking external tools (\textit{e.g.}, application programming interfaces (APIs)) to support final decision execution. 
    % This modular and tool-augmented structure enables it to handle complex tasks with higher efficiency and precision.
\end{itemize}

\subsubsection{Emergence of Agentic AI}
\par The emergence of agentic AI evolves through several stages of AI development, with increasing autonomy and intelligence. In the following, we elaborate on the key developments of AI agent systems to clarify the differences between the agentic AI system and other AI agent systems.
\par \textit{Symbolic/Rule-based Agent}: Conventional rule-based agent relies on predefined rules. In this case, it can only perform specific tasks in static and structured scenarios due to insufficient adaptability and flexibility.
\par \textit{Machine Learning (ML)-based Agent}: Conventional ML-based agent extracts patterns from datasets and performs predictions, which indicates agent systems shift from rule-driven to data-driven. However, it typically requires well-labeled data, exhibits limited adaptability and lacks generalization.
    % \item \textit{DRL-based Agent}: DRL combines ML and deep learning, which learns from the interaction process with the environment and improves the strategy based on the environment feedback. Although DRL agent shows adaptability in dynamic scenarios, it is typically effective in trained domains and lacks generalization.
\par \textit{LLM-based Agent}: LLM-based agent is built upon LLM (\textit{e.g.}, GPT-4) that learns from massive-scale corpora to acquire extensive knowledge, which possesses contextual understanding, reasoning, and generation capabilities. However, LLM-based agent suffers from limited memory and typically lacks proactive environmental awareness.
\par \textit{Agentic AI}: Agentic AI based on the perception-reasoning-action loop integrates autonomy, contextual memory, explicit reasoning, and modular collaboration, enabling long-term planning and proactive decision-making while dynamically adjusting its strategies through active perception of environmental changes. Notably, agentic AI can incorporate tools and multiple heterogeneous models (\textit{e.g.}, LLM and DRL), thereby performing diverse and high-complexity tasks through coordinated operation without frequent retraining.

\subsection{Architectures of ISAC}
\par ISAC systems are typically divided into two architectures, \textit{i.e.}, radar-communication coexistence (RCC) architecture and dual-functional radar and communication (DFRC) architecture. 

\par \textit{RCC Architecture}: In the RCC architecture, sensing and communication systems are physically separated and collaborate through information exchange. This design provides high flexibility and compatibility. However, frequent information exchange results in low resource utilization and high coordination overhead.

\par \textit{DFRC Architecture}: The DFRC architecture integrates sensing and communication within a unified hardware platform. By sharing spectrum and hardware resources, it achieves high resource utilization and low deployment cost. However, the different requirements of sensing and communication introduce inherent performance trade-offs, making joint optimization challenging.

\par Although these two architectures differ in system design and implementation, both require efficient resource management and adaptive decision making to balance sensing and communication performance. As network environments become increasingly dynamic, optimization variables become highly coupled. In this case, agentic AI provides a promising solution by enabling autonomous reasoning, task decomposition, and adaptive decision making. Therefore, the following section reviews recent advances in applying agentic AI to ISAC systems and discusses how agentic AI can improve sensing and communication performance under diverse operating conditions.

\section{Emerging Approaches for ISAC}
\label{Different Optimization Approaches for ISAC}
\par In this section, we review the existing methods and emerging methods in ISAC applications.

\subsection{Existing Methods}
% \par In this part, we introduce the conventional optimization methods that are widely used to solve ISAC optimization problems.

\subsubsection{SCA Methods}
\par The authors in~\cite{Nassar2024} proposed an alternative optimization-based method to maximize the communication rate and sensing power by jointly optimizing the active beamforming matrix, power allocation factor, and reconfigurable intelligent surface (RIS) phase shifts in a hybrid RIS-assisted ISAC system. Specifically, the authors introduced the SCA method by using auxiliary variables and Taylor's approximation to transform the non-convex optimization variables and constraints to those with convex, thereby enabling the CVX tool to iteratively optimize the three optimization variables. Simulation results demonstrated that the proposed method effectively improves the communication rate and sensing power, and it outperforms other comparison methods.

\subsubsection{Game Theory}
\par The authors in~\cite{Liu2024} studied an ISAC system in the presence of a jammer, where the jammer exploits its sensing capabilities to carry out precise attacks. In this case, the authors considered a base station (BS) as the leader and the jammer as the follower and proposed a Bayesian Stackelberg game model. Specifically, the authors derived a closed-form solution to obtain the optimal jamming power in the follower subgame and calculate the optimal beamforming matrix in the leader subgame by using semidefinite relaxation and Gaussian randomization methods. During simulation, the authors explored the impact of channel uncertainty and observation error on the performance of the proposed method, and the results showed that the adopted Stackelberg equilibrium is superior to the Nash equilibrium.

\subsubsection{DRL Methods}
\par The authors in~\cite{Xie2025} aimed to maximize the communication rate and sensing rate of a UAV-carried intelligent reflecting surface (IRS)-assisted ISAC system by jointly optimizing the active beamforming matrix, UAV trajectory, and IRS phase shifts. To solve the optimization problem with dynamic characteristic, the authors proposed an improved DRL-based method. Specifically, the authors utilized the diffusion model and prioritized experience replay (PER) to improve environment analysis capabilities and learning efficiency of the deep deterministic policy gradient (DDPG). Simulation results demonstrated that the proposed DRL-based method outperforms other comparison methods in terms of communication rate and sensing rate.

\par Despite the remarkable achievements of conventional optimization methods, they still suffer from several limitations. Specifically, SCA and game-theoretic methods heavily rely on accurate system knowledge and often require iterative optimization procedures when multiple variables are involved, while DRL methods are highly dependent on reward design and training environments and typically require retraining when the environment changes.

\begin{figure*}
    \centering
    \includegraphics[width=0.9\linewidth]{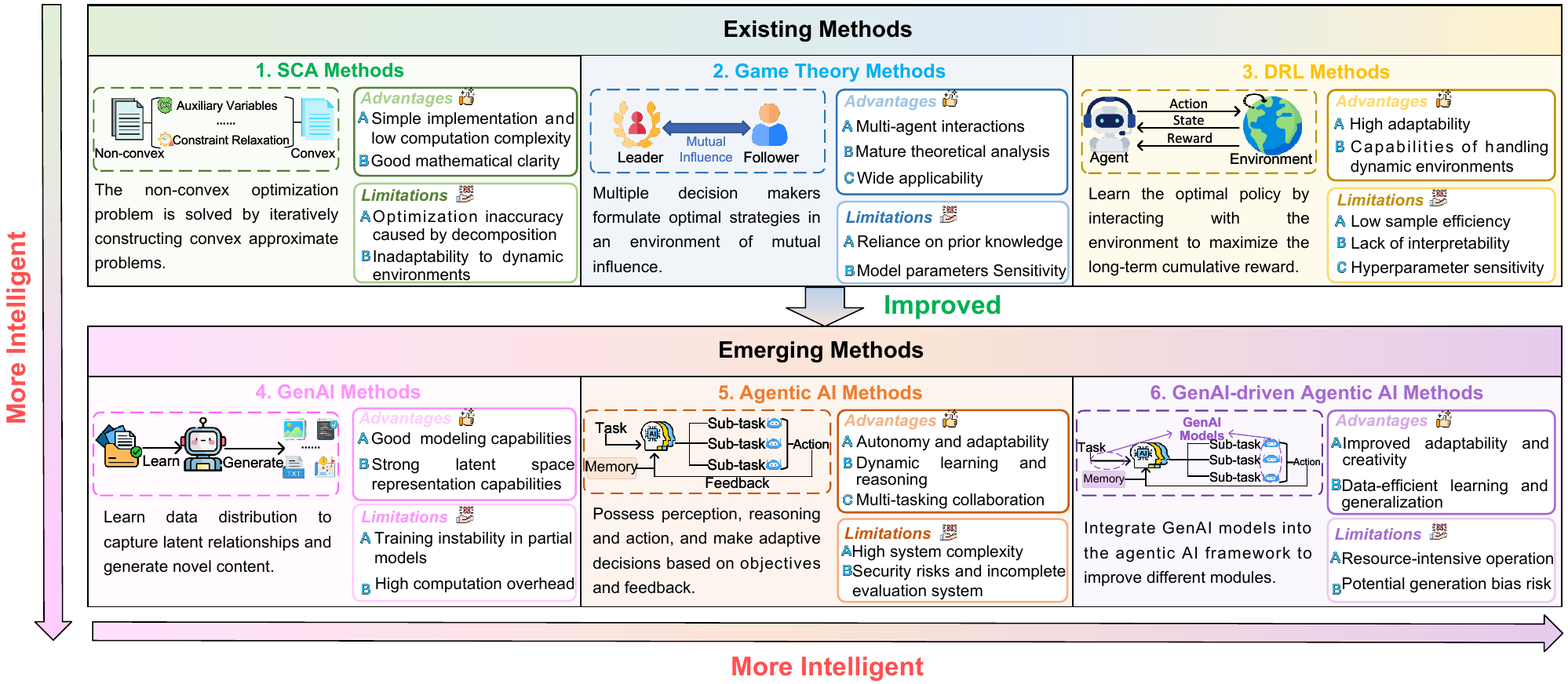}
    \caption{Overview of existing methods and emerging methods for ISAC applications. Optimization methods are developing in a more intelligent direction.}
    \label{fig:Existing Methods and Emerging Methods}
\end{figure*}

\subsection{Emerging Methods}

\subsubsection{Agentic AI Methods}
\par Agentic AI exhibits powerful autonomy and intelligence, which enables ISAC systems to evolve from passive response systems to proactive decision-making systems. Specifically, agentic AI-based ISAC systems can automatically decompose complex and high-level ISAC tasks into multiple low-level subtasks and assign them to appropriate agents, thereby improving the efficiency of solving complex ISAC problems. Moreover, agentic AI-based ISAC systems possess strong memory and self-reflection capabilities, which enables them to adjust their strategies and initiate actions based on historical experiences and feedback from previous decisions. This continuous self-improvement process improves the reasoning and decision accuracy of the involved agents over time. In this case, using agentic AI methods to improve the performance of ISAC systems has become a promising direction that is attracting widespread attention.
% \textcolor{blue}{For instance, an agentic AI-based ISAC system can employ a spectrum knowledge map as persistent memory and use hierarchical task decomposition modules to manageable subtasks based on ISAC objectives. Specifically, the high-level agents are capable of analyzing historical spectrum patterns and environment feedback to generate context-aware subtasks for the low-level agents through a reflection mechanism, which can be implemented by a self-supervised error diagnosis module that compares predicted and observed sensing or communication performance. Then, the low-level agents then perform these subtasks using DRL or other predictive control modules to adjust optimization variables such as transmit power, beamforming matrices, and bandwidth allocation, thereby achieving efficient, high-quality, and automated processing of complex tasks.} In this case, using agentic AI methods to improve the performance of ISAC systems has become a promising direction that is attracting widespread attention.

\par For example, the authors in~\cite{Wang2025} proposed an agentic AI architecture for wireless communications, which can be further extended to ISAC applications. Specifically, they designed an autonomous high-level intent analysis system that maps natural language-based intentions into concrete wireless control actions. First, the proposed framework processes high-level natural-language intents through the Sentence-BERT embedding and K-Means clustering to map user requests into manageable task domains. Based on the classified intent and real-time system state, the DRL agent dynamically selects the most suitable CoT reasoning module. Notably, each selected module provides task-aligned CoT guidance, enabling the LLM to conduct structured step-by-step reasoning on domain-specific processes such as communication modeling, constraint handling, and power-control optimization. This modular CoT design decomposes complex objectives into intermediate reasoning steps, enhancing interpretability and robustness. Then, the natural-language strategies generated by the LLM are converted into executable network control commands through a neural semantic parser. Finally, the real-time feedback (\textit{e.g.}, interference conditions and achieved throughput) is used to continuously improve LLM reasoning modules. Simulation results demonstrate that the proposed CoT module-based wireless system achieves significant improvements in both communication performance and reasoning quality compared to the non-CoT module-based system.

% \par In this agentic AI-based wireless system, multiple pre-trained LLM-based reasoning agents are deployed, each specializing in different wireless tasks. Moreover, several CoT modules are defined within the system, each equipped with distinct prompt templates to guide LLM agents in performing different step-by-step reasoning methods. During operation, the system first analyzes and clusters the high-level intentions obtained from the environment by using a clustering algorithm to select the most appropriate LLM agent. Subsequently, the system utilizes a DRL algorithm to determine the most suitable CoT module based on the clustering results and environmental states, thereby guiding the selected LLM agent to generate reasonable and interpretable wireless control actions. Finally, the system updates the strategies of both the LLM agents and DRL agent through reward feedback. Simulation results demonstrate that the proposed CoT module-based wireless system achieves significant improvements in both communication performance and reasoning quality compared to the non-CoT module-based system.

\par As can be seen, the agentic AI-based ISAC systems can exhibit higher autonomy and stronger generalization ability due to the advanced reasoning mechanisms and massive professional knowledge, which enables them to flexibly handle various heterogeneous tasks and adapt to dynamic and uncertain environments.

% \par Moreover, the authors in [] proposed an agentic AI-based multimodal ISAC framework, which consists of four layers. First, in the multimodal information processing stage, the framework employs a set of pretrained encoders to process the sensed multimodal data and extract semantic features across different modalities. Second, in the multimodal sensing and communication data alignment stage, the framework utilizes an input projection layer to map features from different modalities into a unified representation space, while using a multi-head attention mechanism dynamically to associate textual instructions with communication signal states, enabling the framework to consider both communication context and sensing information when generating control and decision commands. Third, in the multimodal generation empowered downstream task stage, the framework performs prediction and decision based on multimodal inputs to accomplish tasks. Finally, in the multimodal instruction optimization stage, the framework constructs a multimodal instruction tuning dataset and integrates optimization algorithms, allowing the framework to adaptively adjust its decisions based on the environmental feedback in complex and dynamic ISAC scenarios.

\subsubsection{GenAI-driven Agentic AI Methods}
\par Benefiting from the powerful analytical and generative capabilities of GenAI models, the reasoning and decision-making processes of GenAI-driven agentic AI systems can be further enhanced. Currently, since there is almost no specialized research on GenAI-driven agentic AI in ISAC optimization, we introduce the potential applications of these four representative GenAI models in agentic AI-based ISAC systems in the following. 
% In particular, different GenAI models, including generative adversarial networks (GANs), variational autoencoders (VAEs), diffusion models, and Transformers, exhibit unique advantages in various perspectives such as data generation, data reconstruction, denoising, and cross-modal semantic understanding, which provides more comprehensive intelligent support for agentic AI-based ISAC systems.

\par \textit{\textbf{Generative Adversarial Network (GAN)-driven Agentic AI Systems-Enhancing Sample Generation and Policy Robustness}}: The GAN can be used in the reasoning module of the agentic AI system to improve the policy robustness~\cite{Gui2023}. Specifically, the generator learns the latent structure of multimodal perceptual data (such as radar echoes and image information) and generates realistic samples to support the policy training of agents. Meanwhile, the discriminator evaluates the distributional consistency between the generated samples and real observations, which provides feedback on the sample authenticity to the agents. Through this adversarial mechanism, GAN-driven agentic AI can achieve self-learning and policy stabilization even in ISAC scenarios with insufficient real samples. For instance, it is typically challenging to detect low-radar-cross-section objects in ISAC scenarios with rare weather conditions (\textit{e.g.}, fog and rain) due to the lack of real radar samples, which increases detection difficulty. In this case, the GAN serves as a data generator, producing synthetic radar data based on environmental metadata such as fog density, vehicle speed, and multipath intensity. In practical application, when the agentic AI assesses that the uncertainty of the current detection result is high, it autonomously uses the GAN module to generate conditionally matched samples and uses these samples to fine-tune its internal detection model.
% In this way, agentic AI actively expands its data support with GAN-generated data rather than relying solely on limited real-world samples, thereby improving perception reliability and policy robustness in real time.

% For instance, it is typically challenging to detect low radar cross-section objects in ISAC scenarios with rare weather conditions (\textit{e.g.}, fog and rain) due to the lack of real radar samples cause by the increased detection difficulty. In this case, the GAN acts as a data generator to generate synthetic radar data based on environmental metadata such as fog density, vehicle speed, and multipath intensity. In practical application, when the agentic AI assesses that the uncertainty of the current detection result is high, it autonomously uses the GAN module to generate conditionally matched samples and uses these samples to fine-tune its internal detection model. In this way, agentic AI actively expands its data support with GAN-generated data rather than relying solely on limited real-world samples, thereby improving perception reliability and policy robustness in real time.
    
\par \textit{\textbf{Variational Autoencoder (VAE)-driven Agentic AI Systems-Enhancing Implicit Feature Modeling and Signal Reconstruction}}: The VAE can be integrated into the perception module of the agentic AI system to achieve signal reconstruction~\cite{Girin2021}. Specifically, the encoder extracts low-dimensional latent representations from the original complex signals, while the decoder reconstructs signals based on these representations to capture key features. As such, agentic AI equipped with a well-trained VAE can autonomously reconstruct high-confidence signals based on the captured features, even under conditions of low echo energy and sparse sampling scenarios. For instance, in ISAC scenarios involving autonomous driving, radar sampling is typically limited by latency and hardware resources. In this case, the VAE maps sparse noise echoes to a compact latent space and reconstructs high-fidelity echo signals. Then, agentic AI leverages this highly reliable reconstruction result to extract relevant target features (\textit{e.g.}, reflection intensity curves and multipath delays) and perform real-time decisions, such as switching to a narrow beam for precise tracking or enabling additional frequency bands to improve echo quality.
% This enables the agentic AI-based ISAC system to more comprehensively and accurately analyze the current environment, thereby supporting subsequent tasks
% \par For instance, in ISAC scenarios involving autonomous driving, radar sampling is typically limited by latency and hardware resources. In this case, the VAE maps sparse noise echoes to a compact latent space and reconstructs high-fidelity echo signals. Then, agentic AI leverages this highly reliable reconstruction result to extract relevant target features (\textit{e.g.}, reflection intensity curves and multipath delays) and perform real-time decisions, such as switching to a narrow beam for precise tracking or enabling additional frequency bands to improve echo quality. As such, the VAE-driven agentic AI enables reliable perception and control under challenging ISAC environments.

\par \textit{\textbf{Diffusion Model-driven Agentic AI Systems-Enhancing Data Denoising}}: The diffusion model can be integrated into the perception module of the agentic AI system to improve the quality of perception data~\cite{Yang2024}. Specifically, the forward process gradually adds controlled noise to the perceptual data, while the reverse process iteratively denoises and reconstructs clean samples from the noisy data. Considering that ISAC data is typically noisy, this iterative training mechanism enables the agents to learn the underlying distribution of perceptual data, thereby effectively removing both the artificially added noise and inherent noise present in the original perceptual data~\cite{11177562}. For instance, in an urban ISAC monitoring scenario, radar echoes and communication signals typically contain significant noise due to dense traffic, building reflections, and intermittent interference. In this case, the diffusion model learns to recover the original signal and echo distribution from highly contaminated observations through forward-reverse diffusion processes across multiple scenarios offline. In actual operation, when the agentic AI receives highly noisy sensing or communication measurements, it invokes the trained diffusion model to iteratively denoise, thereby reconstructing clearer and more reliable data, such as target echoes and channel state estimates. Based on these clear representations, the agentic AI subsequently performs higher-order inference.

% \par For instance, radar echoes and communication signals in urban ISAC monitoring missions typically contain significant noise due to dense traffic, building reflections, and intermittent interference from uncoordinated equipment. In this case, the diffusion model learns to recover the original signal and echo distribution from highly contaminated observations through forward-reverse diffusion processes across multiple scenarios offline. In actual operation, when the agentic AI receives highly noisy sensing or communication measurements, it invokes the trained diffusion model to iteratively denoise, thereby reconstructing clearer and more reliable data, such as target echoes and channel state estimates. Based on these clear representations, the agentic AI subsequently performs higher-order inference. As such, the diffusion model-driven agentic AI can improve the robustness and reliability of data representation at the perception stage, thereby improving the decision accuracy of downstream tasks.
    
\par \textit{\textbf{Transformer-driven Agentic AI Systems-Enhancing Cross-modal Fusion and Long-term Reasoning}}: The Transformer demonstrates impressive performance in multimodal alignment and spatiotemporal feature modeling due to its powerful attention mechanism and sequence modeling capabilities~\cite{Xu2023}. Therefore, the Transformer can be integrated into the perception module of agentic AI to achieve cross-modal feature fusion and achieve global modeling of dynamic environments through the multi-head attention mechanism. Moreover, the Transformer can also be integrated into the reasoning module, where its powerful long-term dependency modeling capabilities enable agents to perform long-term strategy planning. For instance, UAVs equipped with radar sensors and communication transceivers are deployed to perform ISAC tasks in post-disaster environments. Specifically, the Transformer processes multimodal temporal inputs (\textit{e.g.}, thermal video and radar maps) by aligning their feature sequences via the attention mechanism to extract stable temporal patterns. Based on these multimodal embeddings, the high-level agent evaluates the area situation, such as link reliability trends and flight risks, while low-level agents generate executable actions such as adjusting trajectory and beam directions. This architecture ensures stable environmental sensing and reliable information relay in highly dynamic post-disaster scenarios.
% \par For instance, UAVs equipped with radar sensors and communication transceivers are often deployed to perform ISAC tasks in post-disaster environments. In this case, the Transformer processes multimodal temporal inputs (\textit{e.g.}, high-resolution thermal video, radar maps, and real-time channel state information) by aligning their feature sequences through the attention mechanism and extracting stable temporal patterns and correlations. Then, these multimodal temporal embeddings are provided to the agentic AI. Specifically, the high-level agent evaluates area situation such as link reliability trends and potential flight risks, while the low-level agents generate executable actions such as adjusting trajectory and beam directions, thereby enabling stable environmental sensing and reliable information relay in highly dynamic post-disaster scenarios.

\subsection{Lesson Learned}

\par The above analysis shows that agentic AI methods exhibit strong autonomy and intelligence, thereby overcoming the limitations of conventional ISAC optimization methods~\cite{Wang2025}. In particular, GenAI-driven agentic AI integrates GenAI models into different modules to improve environment analysis, signal enhancement, and strategy optimization. Therefore, different GenAI models can be flexibly incorporated according to the requirements of specific ISAC scenarios. Fig.~\ref{fig:Existing Methods and Emerging Methods} illustrates the evolution of ISAC optimization methods toward greater autonomy and intelligence.

\begin{figure*}
    \centering
    \includegraphics[width=0.9\linewidth]{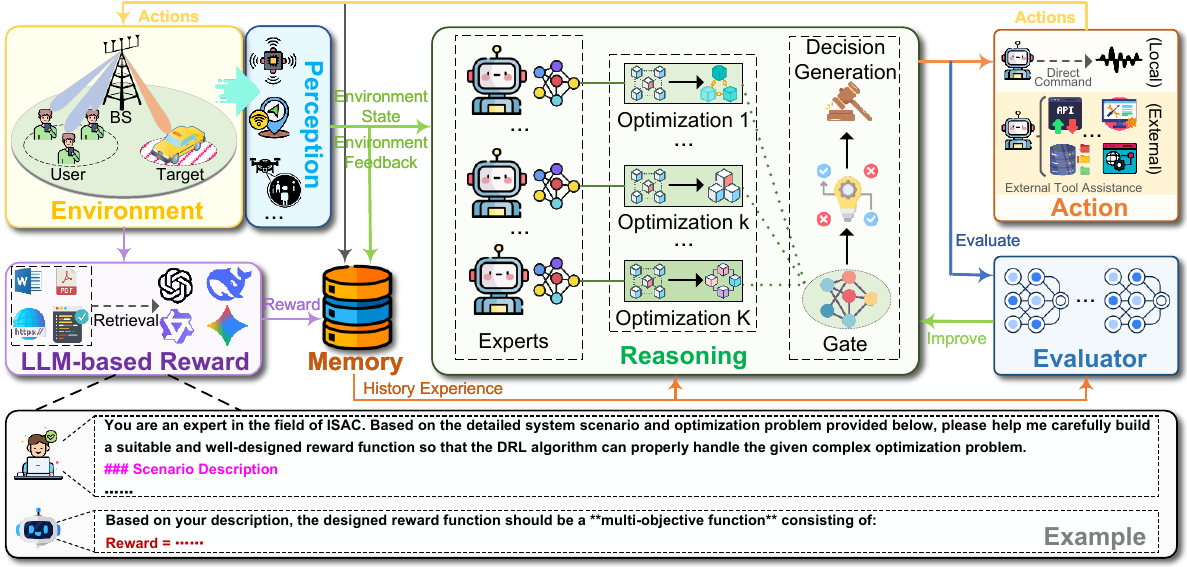}
    \caption{The proposed agentic ISAC framework.}
    \label{fig:Agentic AI Framework}
\end{figure*}

\section{Case Study: Agentic AI for ISAC}

\par In this section, we propose an agentic ISAC framework.

\subsection{Agentic ISAC Framework}

\par We propose an agentic ISAC framework built upon the DRL algorithm, further integrating LLM, GenAI model, and mixture of experts (MoE). Specifically, the LLM utilizes its general knowledge to enable automatic reward function design. The GenAI model improves the environment state analysis of the DRL algorithm via its powerful modeling capability, while MoE significantly enhances its robustness. As shown in Fig.~\ref{fig:Agentic AI Framework}, the framework operates via the perception-reasoning-action loop, comprising perception, reasoning, action, reward, evaluator, and memory modules.
\par \textit{\textbf{Environment Perception}}: In ISAC environments, a large number of sensors are typically deployed to capture diverse critical environmental data, such as GPS-based positioning information and radar-based target perception information. These multimodal data are integrated to form a complete environment state, which serves as the foundation for the agents to fully analyze the surrounding conditions and make informed decisions.
    
\par \textit{\textbf{Reasoning and Planning}}: Based on the observed environment state, the agents perform reasoning and planning through the Transformer-based MoE model. Specifically, the MoE consists of multiple experts, each specialized in handling different tasks, and a gating network that selects the most relevant experts according to the environment state to make decisions. Moreover, considering that each decision has a profound impact on subsequent decisions in ISAC systems, Transformer is integrated into the MoE model to capture this temporal dependency. Specifically, the attention mechanism of the Transformer can adaptively allocate weights based on the relevance among different time steps, thereby highlighting the most critical historical information for the current decision-making and maintaining stable dependency modeling capabilities in long sequences. In this way, the agentic ISAC system can achieve more comprehensive decisions by aggregating the outputs of multiple experts within the reasoning module.
    
\par \textit{\textbf{Action Execution}}: The decisions (\textit{i.e.}, commands) generated by the reasoning module can be executed in two modes. The first is direct command control, where the commands interact directly with the physical environment. The second involves external tool assistance, where the commands are processed through external programs or APIs before interacting with the physical world. In particular, in scenarios that integrate multiple tools, some existing tool engineering methods~\cite{Liu2026} can be integrated into this module to help the framework more intelligently decide when to invoke which tool. Notably, in the agentic ISAC system, decisions can be executed through hybrid modes, where some are performed directly while others are carried out with the assistance of external tools.
    
\par \textit{\textbf{Reward Feedback}}: The reward function directly determines the rationality of the optimization process. However, manual reward design requires extensive ISAC expertise, which poses challenges for newcomers. To address this issue, LLMs can automatically generate reward functions according to system settings and characteristics (\textit{e.g.}, the channel model and location information of components) presented in the prompt. Since LLMs may suffer from hallucinations and lack domain-specific knowledge, the proposed framework incorporates RAG to retrieve relevant knowledge from external sources and assist the reasoning process. It is worth noting that complex ISAC scenarios often involve massive background information, which makes prompt design challenging in such cases. Excessive background information in the prompt may interfere with the reasoning process of the LLM and is unsuitable for scenarios with limited reasoning budgets, whereas insufficient information may reduce reasoning accuracy. To address this issue, the proposed framework can be integrated with the existing wireless context engineering framework~\cite{zhao2026wireless}, which automatically extracts key task-related background information and converts it into structured prompts. As such, the framework can prioritize high-impact cues under limited inference budgets, thereby enabling efficient operation in complex ISAC scenarios over resource-constrained networks.
    
\par \textit{\textbf{Evaluation and Update}}: The evaluation module is responsible for gradually improving the decision accuracy of the reasoning module. Specifically, it continuously monitors and analyzes the decisions generated by the reasoning module, leveraging historical experience with reward feedback to evaluate the current strategy quality of the reasoning module, thereby providing optimization guidance for the reasoning process.
    
\par \textit{\textbf{Memory Storage}}: The memory module stores the generated experiences, which summarize the interactions between the agents and environment. These experiences accumulate over time, forming a rich historical knowledge base from which the agents can learn to improve future decision-making and ultimately improve their performance in dynamic environments.

\par To ensure real-time responsiveness, the proposed framework adopts an offline-training and online-deployment paradigm. Specifically, computationally intensive tasks, including LLM-driven reward design and multi-agent training, are completed offline. During online deployment, the framework only performs low-complexity forward propagation. In particular, the MoE architecture employs sparse activation, where the gating network selects only the most relevant experts for each decision. As such, the online decision overhead is significantly reduced, thereby satisfying the latency requirements of dynamic ISAC scenarios.

\begin{figure}
    \centering
    \includegraphics[width=\linewidth]{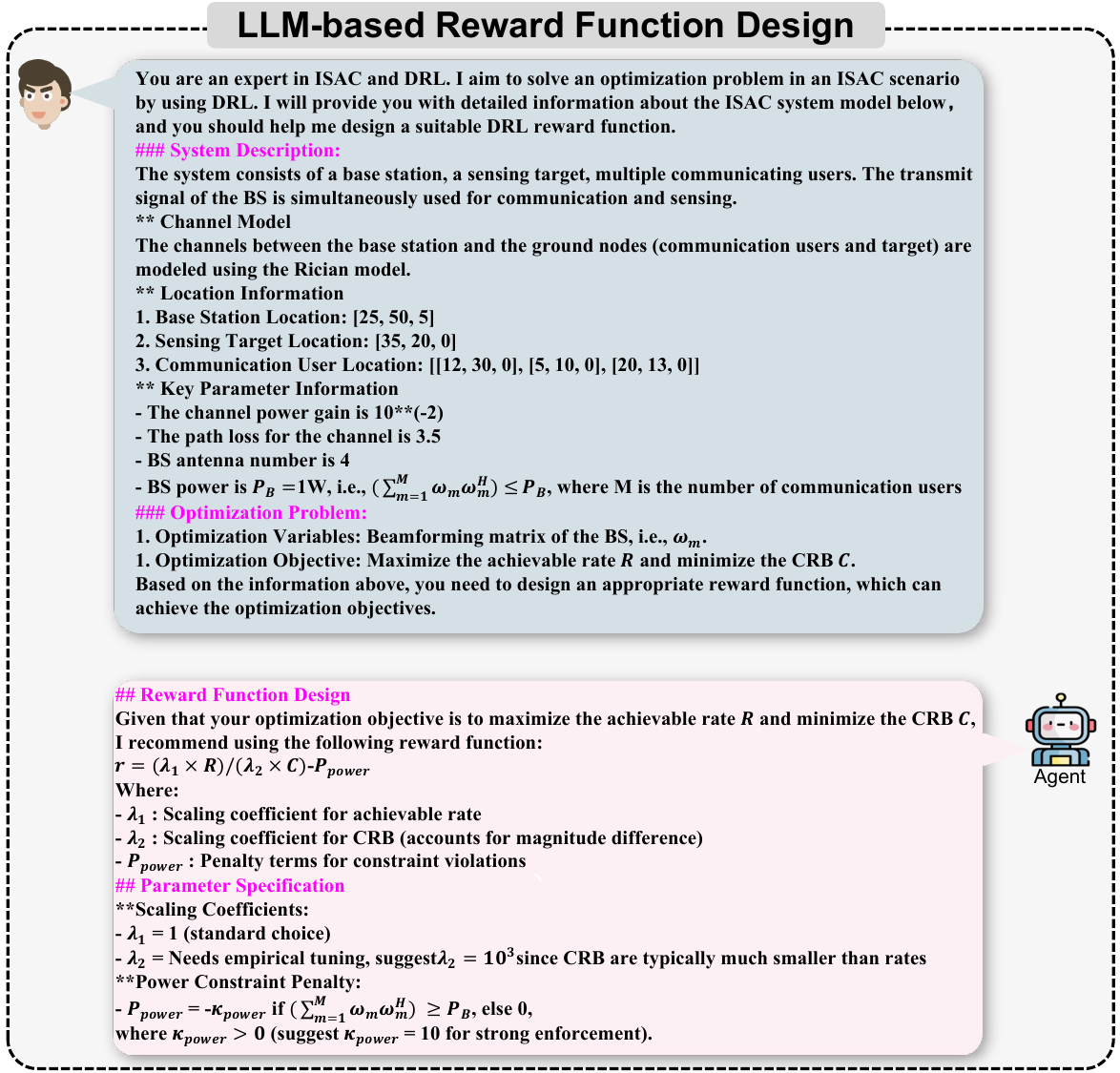}
    \caption{The LLM-designed reward function based on the system model description and formulated optimization problem.}
    \label{fig:agent-human interaction}
\end{figure}

\subsection{Simulation}
\subsubsection{System Model Description} 
\par To evaluate the effectiveness of the proposed agentic ISAC framework, we consider a dual-functional BS-enabled sensing and communication system, which consists of multiple ground users and a target. Specifically, the BS provides communication services to the users while attempting to estimate the location information of the target. To improve the communication quality and positioning accuracy, we aim to maximize the communication rate and minimize the Cram\'{e}r-Rao bound (CRB) by optimizing the active beamforming matrix of the BS, where CRB represents the theoretical lower bound on the accuracy of position parameter estimation.

\subsubsection{Performance Analysis}
\par
\begin{figure}
    \centering
    \includegraphics[width=0.78\linewidth]{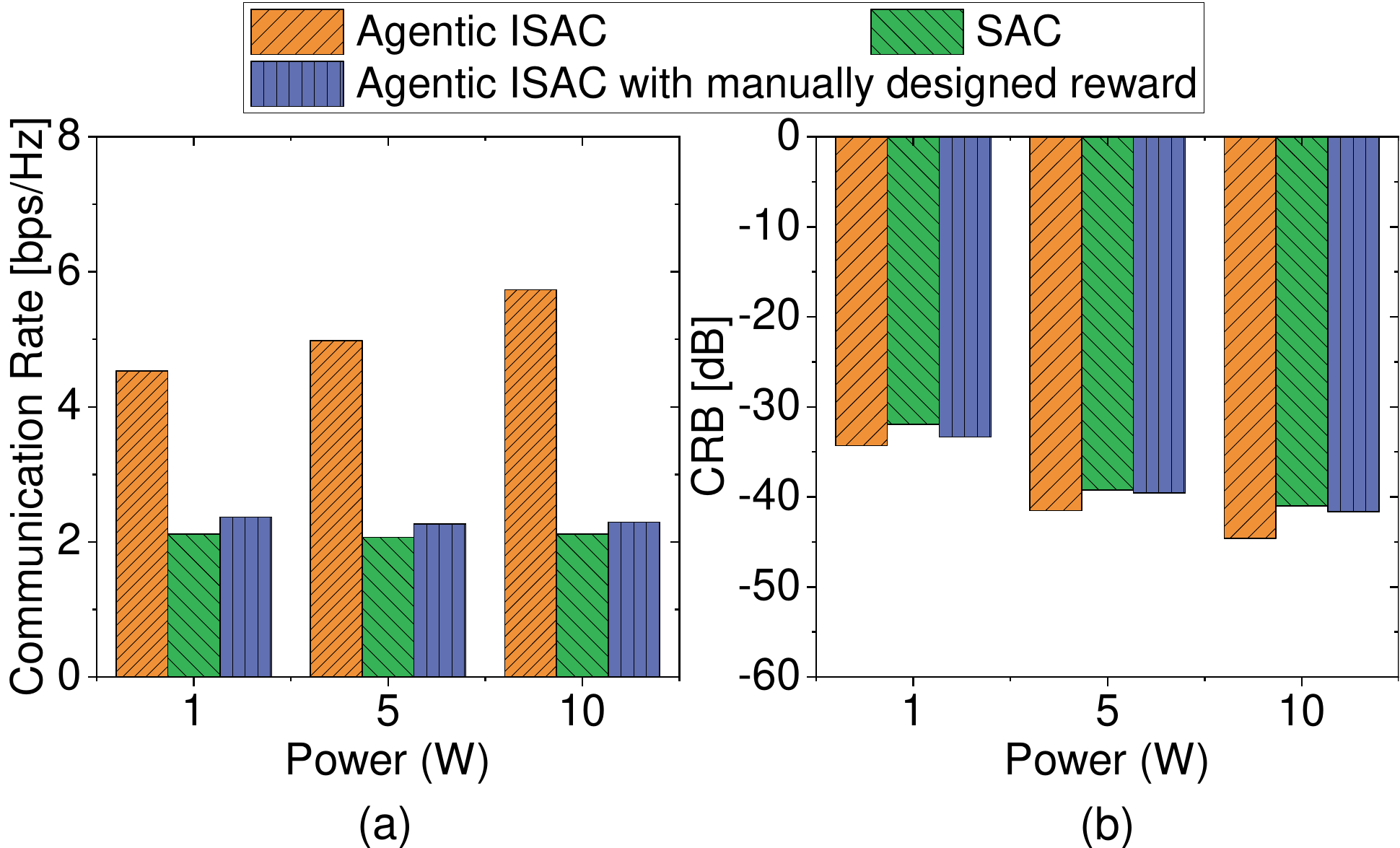}
    \caption{The impact of BS transmit power on different optimization objectives. (a) Communication rate. (b) CRB.}
    \label{fig:optimization values}
\end{figure}

 We adopt the soft actor-critic (SAC) algorithm as the basic DRL algorithm in the proposed agentic ISAC framework. Fig.~\ref{fig:agent-human interaction} shows the interaction process in which the LLM designs the reward function. As can be seen, the LLM is able to generate a well-structured reward function based on the system model configuration and optimization problem provided in the prompt, where the designed reward function satisfies the optimization objectives while considering the transmit power constraint of the BS. Moreover, the LLM explicitly considers the difference in magnitude between the communication rate and CRB, which effectively prevents learning bias during training.  Fig.~\ref{fig:optimization values} shows the communication rate and CRB achieved by agentic AI, SAC, and agentic AI with a manually designed reward function under different BS transmit power settings. As can be seen, the reward function generated by the LLM outperforms the manually designed one, which demonstrates the effectiveness of LLM reasoning in balancing multiple optimization objectives. Moreover, the proposed agentic ISAC framework consistently outperforms SAC due to the strong feature extraction capability of the Transformer architecture and the robust decision-making capability of the MoE-based actor network. In addition, the communication rate increases while the CRB decreases as the BS transmit power increases because a higher transmit power improves the SNR of the links.

\section{Future Directions}
\par \textbf{Secure Agentic AI Framework for ISAC:} Agentic AI frameworks rely on LLMs and knowledge databases for decision making. Since erroneous data may cause incorrect decisions and degrade system performance, ensuring data integrity, confidentiality, and tamper resistance is critical. Therefore, blockchain and differential privacy are promising solutions for secure agentic AI systems.

\par \textbf{Lightweight Agentic AI Framework for ISAC:} Since agentic AI frameworks typically integrate multiple methods to handle different tasks and challenges, including resource-intensive methods such as DRL, GenAI, and LLM. In this case, deploying an agentic AI framework to perform ISAC-related decisions becomes highly challenging in resource-constrained ISAC applications. Therefore, developing lightweight agentic AI for ISAC can improve its usability and deployment efficiency.
% which can be achieved by introducing compact decision modules or applying resource-efficient distillation strategies to reduce computational overhead and maintain stable performance under limited computation resource.

\par \textbf{Cross-Domain Agentic AI Framework for ISAC:} Developing a cross-domain agentic AI framework that integrates knowledge and reasoning mechanisms from different domains can improve the decision-making process of the agentic ISAC system. Specifically, this framework first extracts key features through a cross-domain information fusion module and utilizes a cross-domain transfer mechanism to map the structural information learned in one task to another. 
Moreover, the framework deploys a unified planning module to possess cross-task collaborative abilities in complex environments.

% \par \textbf{Cross-Domain Agentic AI Framework for ISAC:} Current agentic AI frameworks typically rely solely on domain knowledge that is directly related to the task, which may limit their decision-making capabilities. In this case, developing a cross-domain agentic AI framework that integrates knowledge and reasoning mechanisms from different domains can improve the decision-making capabilities of the agentic ISAC system. Specifically, this framework first extracts key features through a cross-domain information fusion module and utilizes a cross-domain transfer mechanism to map the structural information learned in one task to another. 
% Based on this, the framework outputs decisions through a unified planning module, enabling the ISAC system to possess stronger cross-task collaborative abilities in complex environments.

\section{Conclusion}
\par In this paper, we have systematically reviewed how agentic AI can be applied to ISAC systems to enable intelligent and automated decision-making. First, we have traced the evolution of agentic AI and analyzed the characteristics of different ISAC architectures. Second, we have summarized both existing and emerging optimization approaches in ISAC. Third, we have proposed an agentic ISAC framework and have validated its effectiveness through a case study. Specifically, simulation results have demonstrated that the LLM-based reward function has outperformed manually crafted ones. Moreover, the proposed agentic AI framework has achieved an improvement of 131.25\% in communication rate and 5.43\% in CRB, which indicates the effectiveness of the proposed framework due to the MoE architecture and GenAI model. Finally, we have outlined several promising research directions for future studies.
% We hope that this work has provided useful insights and inspiration for further research on agentic AI-driven ISAC optimization.

\normalem

\bibliography{main}

\end{document}